\documentclass[11pt,a4paper]{article}
\usepackage[hyperref]{acl2018}
\usepackage{times}
\usepackage{latexsym}
\usepackage{graphicx}
\usepackage{float}

\usepackage{url}
\usepackage{multirow}
\usepackage{booktabs}
\usepackage{caption}
\DeclareCaptionFont{10pt}{\fontsize{10pt}{12pt}\selectfont}
\usepackage{subcaption}
\usepackage[flushleft]{threeparttable}
\usepackage{todonotes}

\definecolor{DarkRed}{RGB}{130,25,0}

\newcommand{\missing}[1]{\textcolor{red}{#1}}
\newcommand{\ignore}[1]{}
\newcommand{\event}[1]{\textit{\textbf{#1}}}
\newcommand{\idxevent}[3]{\event{e#1:\textrm{\color{#3}#2}}}
\newcommand{\refevent}[2]{\event{e\ref{#1}:#2}}
\newcommand{\rel}[1]{{\em #1}}
\newcommand{\type}[1]{\textsc{#1}}
\newcommand{\anothertype}[1]{\textsl{#1}}

\newcommand\QNfinal[1]{#1}
\newcommand{\best}[1]{\textbf{#1}}
\newcommand{\dataset}{MATRES}
\newcounter{exctr}
\newcommand{\addexctr}{\refstepcounter{exctr}\theexctr}
\newcounter{eventCtr}
\newcommand{\addeventCtr}{\refstepcounter{eventCtr}\theeventCtr}

\aclfinalcopy 
\def\aclpaperid{***} 

\title{A Multi-Axis Annotation Scheme for Event Temporal Relations}

\author{Qiang Ning,$^1$ Hao Wu,$^2$ Dan Roth$^{1,2}$ \\
	Department of Computer Science\\
	$^1$University of Illinois at Urbana-Champaign, Urbana, IL 61801, USA\\
	$^2$University of Pennsylvania, Philadelphia, PA 19104, USA\\
	{\tt \small qning2@illinois.edu,~\{haowu4,danroth\}@seas.upenn.edu}}

\date{}

\begin{document}
\maketitle
\begin{abstract}
	Existing temporal relation (TempRel) annotation schemes often have low inter-annotator agreements (IAA) even between experts, suggesting that the current annotation task needs a better definition.
	This paper proposes a new multi-axis modeling to better capture the temporal structure of events.
	In addition, we identify that event end-points are a major source of confusion in annotation, so we also propose to annotate TempRels based on start-points only.
	A pilot expert annotation effort using the proposed scheme shows significant improvement in IAA from the conventional 60's to 80's  (Cohen's Kappa).
	This better-defined annotation scheme further enables the use of crowdsourcing to alleviate the labor intensity for each annotator.
	We hope that this work can foster more interesting studies towards event understanding.\footnote{\QNfinal{The dataset is publicly available at \url{https://cogcomp.org/page/publication_view/834}.}}
\end{abstract}
\section{Introduction}
Temporal relation (TempRel) extraction is an important task for event understanding, and it has drawn much attention in the natural language processing (NLP) community recently \cite{ULADVP13,ChambersCaMcBe14,LCUMAP15,MSAAVMRUK15,BDSPV15,BSCDPV16,BSPP17,LeeuwenbergMo17,NingFeRo17,NingWuPeRo18,NingYuFaRo18}.

Initiated by TimeBank (TB) \cite{PHSSGSRSDF03}, a number of TempRel datasets have been collected, including but not limited to the verb-clause augmentation to TB \cite{BethardMaKl07}, TempEval1-3 \cite{VGSHKP07,VSCP10,ULADVP13}, TimeBank-Dense (TB-Dense) \cite{CassidyMcChBe14}, EventTimeCorpus \cite{ReimersDeGu16}, and datasets with both temporal and other types of relations (e.g., coreference and causality) such as CaTeRs \cite{CaTeRs} and RED \cite{GormanWrPa16}. 
These datasets were annotated by experts, but most still suffered from low inter-annotator agreements (IAA).
For instance, the IAAs of TB-Dense, RED and THYME-TimeML \cite{SBFPPG14} were only below or near 60\% (given that events are already annotated).
Since a low IAA usually indicates that the task is difficult even for humans (see Examples~\ref{ex:hard cases 1}-\ref{ex:hard cases 3}), the community has been looking into ways to simplify the task, by reducing the label set, and by breaking up the overall, complex task into subtasks (e.g., getting agreement on which event pairs should have a relation, and then what that relation should be) \cite{CaTeRs,GormanWrPa16}.
\QNfinal{In contrast to other existing datasets, \citet{BethardMaKl07} achieved an agreement as high as 90\%, but the scope of its annotation was narrowed down to a very special verb-clause structure.}

\begin{table}[h!]
	\centering\small
	\begin{tabular}{|p{7.5cm}|}
		\hline
		\textbf{(\event{e\ref{ev:restore}}, \event{e\ref{ev:killed}}), (\event{e\ref{ev:showed}}, \event{e\ref{ev:hit}}), and (\event{e\ref{ev:rebuilding}}, \event{e\ref{ev:responded}}): TempRels that are difficult even for humans. Note that only relevant events are highlighted here.}\\
		\hline
		\textbf{Example~\addexctr\label{ex:hard cases 1}:}
		Serbian police tried to eliminate the pro-independence Kosovo Liberation Army and (\idxevent{\addeventCtr\label{ev:restore}}{restore}{black}) order.
		At least 51 people were (\idxevent{\addeventCtr\label{ev:killed}}{killed}{black}) in clashes between Serb police and ethnic Albanians in the troubled region.\\
		\hline
		\textbf{Example~\addexctr\label{ex:hard cases 2}:}
		Service industries (\idxevent{\addeventCtr\label{ev:showed}}{showed}{black}) solid job gains, as did manufacturers, two areas expected to be hardest (\idxevent{\addeventCtr\label{ev:hit}}{hit}{black}) when the effects of the Asian crisis hit the American economy.\\
		\hline
		\textbf{Example~\addexctr\label{ex:hard cases 3}:}
		We will act again if we have evidence he is (\idxevent{\addeventCtr\label{ev:rebuilding}}{rebuilding}{black}) his weapons of mass destruction capabilities, senior officials say. In a bit of television diplomacy, Iraq's deputy foreign minister (\idxevent{\addeventCtr\label{ev:responded}}{responded}{black}) from Baghdad in less than one hour, saying that \dots\\
		\hline
	\end{tabular}
\end{table}

This paper proposes a new approach \QNfinal{to handling these issues in TempRel annotation}.
\textbf{First}, we introduce {\em multi-axis modeling} to represent the temporal structure of events, based on which we anchor events to different semantic axes; only events from the same axis will then be temporally compared (Sec.~\ref{sec:structure}). As explained later, those event pairs in Examples~\ref{ex:hard cases 1}-\ref{ex:hard cases 3} are difficult because they represent different semantic phenomena and belong to different axes.
\textbf{Second}, while we represent an event pair using two time intervals (say, $[t_{start}^1,t_{end}^1]$ and $[t_{start}^2,t_{end}^2]$), we suggest that comparisons involving end-points (e.g., $t_{end}^1$ vs. $t_{end}^2$) are typically more difficult than comparing start-points (i.e., $t_{start}^1$ vs. $t_{start}^2$); we attribute this to the ambiguity of expressing and perceiving durations of events \cite{CollGe11}.
We believe that this is an important consideration, and we propose in Sec.~\ref{sec:split} that TempRel annotation should focus on start-points.
Using the proposed annotation scheme, a pilot study done by experts achieved a high IAA of .84 (Cohen's Kappa) on a subset of TB-Dense, in contrast to the conventional 60's.

In addition to the low IAA issue, TempRel annotation is also known to be labor intensive. 
Our \textbf{third contribution} is that we facilitate, for the first time, the use of  crowdsourcing to collect a new, high quality (under multiple metrics explained later) TempRel dataset. 
We explain how the crowdsourcing quality was controlled and how \rel{vague} relations were handled in Sec.~\ref{sec:annotation}, and present some statistics and the quality of the new dataset in Sec.~\ref{sec:stat}.
A baseline system is also shown to achieve much better performance on the new dataset, when  compared with system performance in the literature (Sec.~\ref{sec:result}).
The paper's results are very encouraging and hopefully, this work would significantly benefit research in this area.
\section{Temporal Structure of Events}
\label{sec:structure}

Given a set of events, one important question in designing the TempRel annotation task is: which pairs of events should have a relation? The answer to it depends on the modeling of the overall temporal structure of events.

\subsection{Motivation}
\label{subsec:motivation}
TimeBank \cite{PHSSGSRSDF03} laid the foundation for many later TempRel corpora, e.g., \cite{BethardMaKl07, ULADVP13, CassidyMcChBe14}.\footnote{EventTimeCorpus \cite{ReimersDeGu16} is based on TimeBank, but aims at anchoring events onto explicit time expressions in each document rather than annotating TempRels between events, which can be a good complementary to other TempRel datasets.}
In TimeBank, the annotators were allowed to label TempRels between any pairs of events. 
This setup models the overall structure of events using {\em a general graph}, which made annotators inadvertently overlook some pairs, resulting in low IAAs and many false negatives.

\begin{table}[h!]
	\centering\small
	\begin{tabular}{|p{7.5cm}|}
		\hline
		\textbf{Example~\addexctr\label{ex:dense scheme}: Dense Annotation Scheme.}\\
		\hline
		Serbian police (\idxevent{\addeventCtr\label{ev:tried}}{tried}{black}) to (\idxevent{\addeventCtr\label{ev:eliminate}}{eliminate}{black}) the pro-independence Kosovo Liberation Army and (\refevent{ev:restore}{restore}) order.
		At least 51 people were (\refevent{ev:killed}{killed}) in clashes between Serb police and ethnic Albanians in the troubled region.\\
		\hline
		\textbf{Given 4 \type{Non-Generic} events above, the dense scheme presents 6 pairs to annotators one by one: (\event{e\ref{ev:tried}}, \event{e\ref{ev:eliminate}}), (\event{e\ref{ev:tried}}, \event{e\ref{ev:restore}}), (\event{e\ref{ev:tried}}, \event{e\ref{ev:killed}}), (\event{e\ref{ev:eliminate}}, \event{e\ref{ev:restore}}), (\event{e\ref{ev:eliminate}}, \event{e\ref{ev:killed}}), and (\event{e\ref{ev:restore}}, \event{e\ref{ev:killed}}). Apparently, not all pairs are well-defined, e.g., (\event{e\ref{ev:eliminate}}, \event{e\ref{ev:killed}}) and (\event{e\ref{ev:restore}}, \event{e\ref{ev:killed}}), but annotators are forced to label all of them.}\\
		\hline
		
		\ignore{\event{DCT:02/27/1998}:The mayor of Moscow has (\idxevent{\addeventCtr\label{ev:allocated}}{allocated}{black}) funds to (\idxevent{\addeventCtr\label{ev:build}}{build}{black}) a museum in honor of Mikhail Kalashnikov, (\idxevent{\addeventCtr\label{ev:said}}{said}{black}) the culture minister. Mikhail Kalashnikov (\idxevent{\addeventCtr\label{ev:designed}}{designed}{black}) the AK-47 automatic rifle.
			\\
			{\em
				$C_4^2$ pairs between \event{e\ref{ev:allocated}}-\event{e\ref{ev:designed}} will be annotated one by one in this window.}\\
			\hline
			\event{DCT:02/13/1998}:``He (\idxevent{\addeventCtr\label{ev:knows}}{knows}{black}) that we will (\idxevent{\addeventCtr\label{ev:act}}{act}{black}) if we (\idxevent{\addeventCtr\label{ev:have}}{have}{black}) evidence that he is (\idxevent{\addeventCtr\label{ev:building}}{building}{black}) his weapons of mass destruction capabilities.'' In less than an hour, the deputy foreign minister (\idxevent{\addeventCtr\label{ev:responded}}{responded}{black}) from Baghdad, (\idxevent{\addeventCtr\label{ev:saying}}{saying}{black}) \dots\\
			{\em $C_6^2$ pairs between \event{e\ref{ev:knows}}-\event{e\ref{ev:saying}} will be annotated one by one in this window.}\\
			\hline}
	\end{tabular}
\end{table}

To address this issue, \citet{CassidyMcChBe14} proposed a dense annotation scheme, TB-Dense, which annotates all event pairs within a sliding, two-sentence window (see Example~\ref{ex:dense scheme}).
It requires all TempRels between \type{Generic}\footnote{For example, {\em lions eat meat} is \type{Generic}.} and \type{Non-Generic} events to be labeled as \rel{vague}, which conceptually models the overall structure by {\em two disjoint time-axes}: one for the \type{Non-Generic} and the other one for the \type{Generic}.

However, as shown by Examples~\ref{ex:hard cases 1}-\ref{ex:hard cases 3} in which the highlighted events are \type{Non-Generic}, the TempRels may still be ill-defined: 
In Example~\ref{ex:hard cases 1}, Serbian police tried to restore order but ended up with conflicts. 
It is reasonable to argue that the attempt to \refevent{ev:restore}{restore} order happened \rel{before} the conflict where 51 people were \refevent{ev:killed}{killed}; or, 51 people had been \event{killed} but order had not been \event{restored} yet, so \refevent{ev:restore}{restore} is \rel{after} \refevent{ev:killed}{killed}.
Similarly, in Example~\ref{ex:hard cases 2}, service industries and manufacturers were originally expected to be hardest \refevent{ev:hit}{hit} but actually \refevent{ev:showed}{showed} gains, so \refevent{ev:hit}{hit} is \rel{before} \refevent{ev:showed}{showed}; however, one can also argue that the two areas had \event{showed} gains but had not been \event{hit}, so \refevent{ev:hit}{hit} is \rel{after} \refevent{ev:showed}{showed}.
Again, \refevent{ev:rebuilding}{rebuilding} is a hypothetical event: ``we will act if \event{rebuilding} is true''. Readers do not know for sure if ``he is already rebuilding weapons but we have no evidence'', or ``he will be building weapons in the future'', so annotators may disagree on the relation between \refevent{ev:rebuilding}{rebuilding} and \refevent{ev:responded}{responded}.
Despite, importantly, minimizing missing annotations, the current dense scheme forces annotators to label many such ill-defined pairs, resulting in low IAA.

\subsection{Multi-Axis Modeling}
\label{subsec:anchor}

Arguably, an ideal annotator may figure out the above ambiguity by him/herself and mark them as \rel{vague}, but it is not a feasible requirement for all annotators to stay clear-headed for hours; let alone crowdsourcers.
What makes things worse is that, after annotators spend a long time figuring out these difficult cases, whether they disagree with each other or agree on the vagueness, the final decisions for such cases will still be {\em vague}.

As another way to handle this dilemma, TB-Dense resorted to a 80\% confidence rule: annotators were allowed to choose a label if one is 80\% sure that it was the writer's intent. 
However, as pointed out by TB-Dense, annotators are likely to have rather different understandings of 80\% confidence and it will still end up with disagreements.

In contrast to these annotation difficulties, humans can easily grasp the meaning of news articles, implying a potential gap between the difficulty of the annotation task and the one of understanding the actual meaning of the text.
In Examples~\ref{ex:hard cases 1}-\ref{ex:hard cases 3}, the writers did not intend to explain the TempRels between those pairs, and the original annotators of TimeBank\footnote{Recall that they were given the entire article and only salient relations would be annotated.} did not label relations between those pairs either, which indicates that both writers and readers did not think the TempRels between these pairs were crucial.
Instead, what is crucial in these examples is that ``Serbian police \event{tried} to restore order but \event{killed} 51 people", that ``two areas were \event{expected} to be hit but \event{showed} gains", and that ``\event{if} he rebuilds weapons \event{then} we will act." To ``\event{restore} order",  to be ``hardest \event{hit}", and ``if he was \event{rebuilding}'' were only the intention of police, the opinion of economists, and the condition to \event{act}, respectively, and whether or not they actually happen is not the focus of those writers.

\begin{table}
	\centering\small
	\begin{tabular}{c|c}
		\hline
		Event Type	&	Category\\
		\hline
		\type{Intention}, \type{Opinion}	&	On an orthogonal axis\\
		\type{Hypothesis}, \type{Generic}	&	On a parallel axis\\
		\type{Negation}	&	Not on any axis\\
		\type{Static}, \type{Recurrent}	&	Other\\
		\hline
	\end{tabular}
	\caption{The interpretation of various  event types \QNfinal{that are not on the main axis} in the proposed multi-axis modeling. The names are rather straightforward; see examples for each in Appendix~\ref{appsec:definition}.}
	\label{tab:nonanchorable}
\end{table}

\begin{figure}[htbp!]
	\includegraphics[width=.55\textwidth]{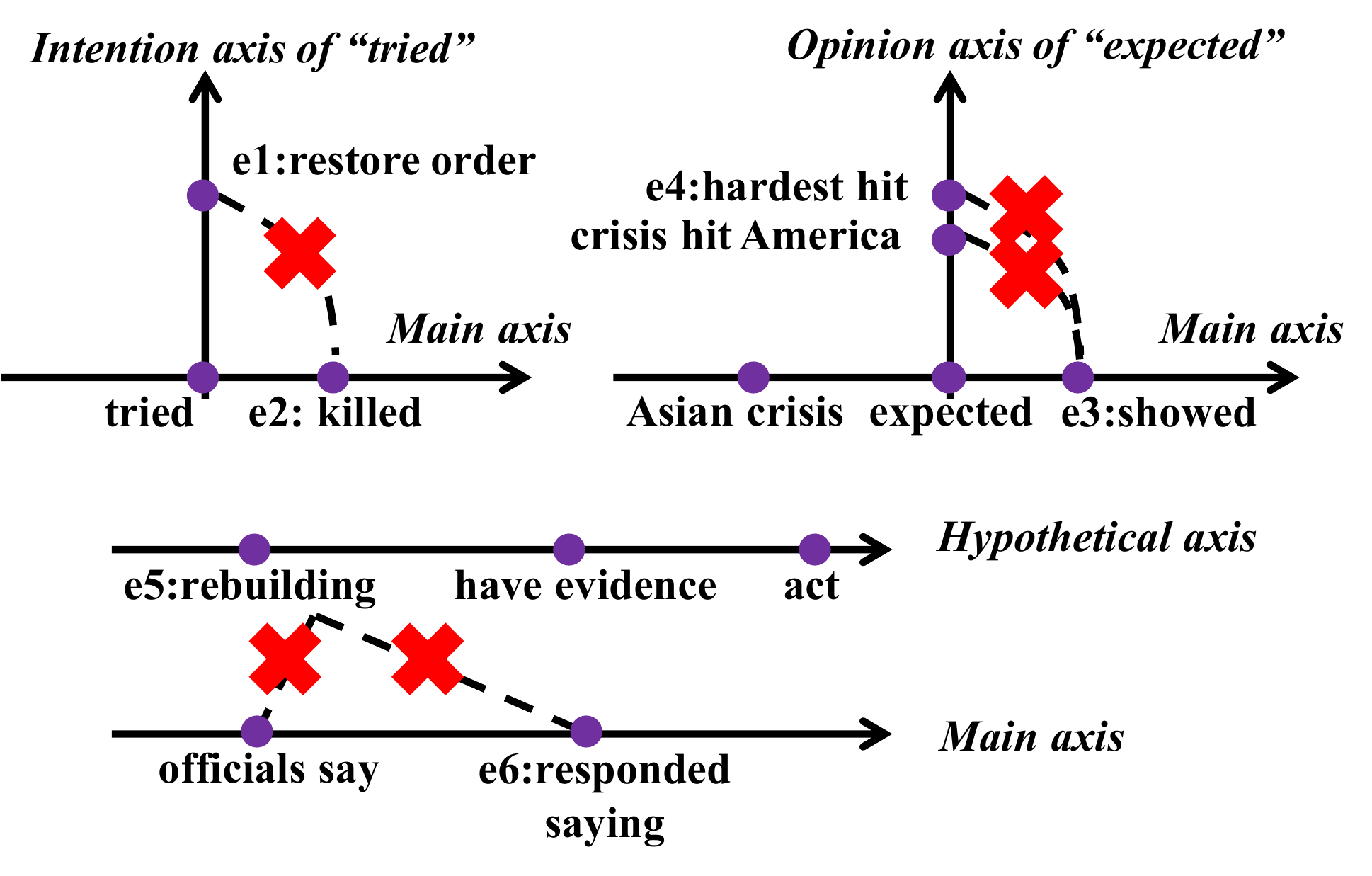}
	\caption{A multi-axis view of Examples~\ref{ex:hard cases 1}-\ref{ex:hard cases 3}. Only events on the same axis are compared.}
	\label{fig:multi axes}
\end{figure}

This discussion suggests that a single axis is too restrictive to represent the complex structure of \type{Non-Generic} events.
Instead, we need a modeling which is more restrictive than a general graph so that annotators can focus on relation annotation (rather than looking for pairs first), but also more flexible than a single axis so that ill-defined relations are not forcibly annotated.
Specifically, we need axes for intentions, opinions, hypotheses, etc. in addition to the main axis of an article. We thus argue for {\em multi-axis modeling}, as defined in Table~\ref{tab:nonanchorable}.
Following the proposed modeling, Examples~\ref{ex:hard cases 1}-\ref{ex:hard cases 3} can be represented as in Fig.~\ref{fig:multi axes}.
This modeling aims at capturing what the author has explicitly expressed and it only asks annotators to look at comparable pairs, rather than forcing them to make decisions on often vaguely defined pairs.

In practice, we annotate one axis at a time: we first classify if an event is anchorable onto a given axis \QNfinal{(this is also called the anchorability annotation step)};
then we annotate every pair of anchorable events \QNfinal{(i.e., the relation annotation step)};
finally, we can move to another axis \QNfinal{and repeat the two steps above}.
\QNfinal{Note that ruling out cross-axis relations is only a strategy we adopt in this paper to separate well-defined relations from ill-defined relations. We do not claim that cross-axis relations are unimportant; instead, as shown in  Fig.~\ref{fig:projection}, we think that cross-axis relations are a different semantic phenomenon that requires additional investigation.}

\subsection{Comparisons with Existing Work}
\QNfinal{There have been other proposals of temporal structure modelings \cite{BDLB06,BeKoMoMa12}, but in general, the semantic phenomena handled in our work are very different and complementary to them. \cite{BDLB06} introduces ``temporal segments'' (a fragment of text that does not exhibit abrupt changes) in the medical domain. Similarly, their temporal segments can also be considered as a special temporal structure modeling. But a key difference is that \cite{BDLB06} only annotates inter-segment relations, ignoring intra-segment ones. Since those segments are usually large chunks of text, the semantics handled in \cite{BDLB06} is in a very coarse granularity (as pointed out by \cite{BDLB06}) and is thus different from ours.}

\QNfinal{\cite{BeKoMoMa12} proposes a tree structure for children's stories, which ``typically have simpler temporal structures'', as they pointed out. Moreover, in their annotation, an event can only be linked to a single nearby event, even if multiple nearby events may exist, whereas we do not have such restrictions.}

\QNfinal{In addition}, some of the semantic phenomena in Table~\ref{tab:nonanchorable} have been discussed in existing work.
Here we compare with them for a better positioning of the proposed scheme.

\subsubsection{Axis Projection}
\QNfinal{TB-Dense handled the incomparability between main-axis events and \type{Hypothesis}/\type{Negation} by {\em treating an event as having occurred} if the event is \type{Hypothesis}/\type{Negation}.\footnote{In the case of Example~\ref{ex:hard cases 3}, it is to treat \event{rebuilding} as actually happened and then link it to \event{responded}.} 
In our multi-axis modeling, the strategy adopted by TB-Dense falls into a more general approach, ``axis projection''. That is, projecting events across different axes to handle the incomparability between any two axes (not limited to \type{Hypothesis}/\type{Negation}).}
Axis projection works well for certain event pairs like \event{Asian crisis} and \refevent{ev:hit}{hardest hit} in Example~\ref{ex:hard cases 2}: as in Fig.~\ref{fig:multi axes}, \event{Asian crisis} is \rel{before} \event{expected}, which is again \rel{before} \refevent{ev:hit}{hardest hit}, so \event{Asian crisis} is \rel{before} \refevent{ev:hit}{hardest hit}.

\QNfinal{Generally, however,} since there is no direct evidence that can guide the projection, annotators may have different projections (imagine projecting \refevent{ev:rebuilding}{rebuilding} onto the main axis: is it in the past or in the future?).
As a result, axis projection requires many specially designed guidelines or strong external knowledge. 
Annotators have to rigidly follow the sometimes counter-intuitive guidelines or ``guess'' a label instead of looking for evidence in the text.

\begin{figure}[htbp!]
	\centering
	\includegraphics[width=.3\textwidth]{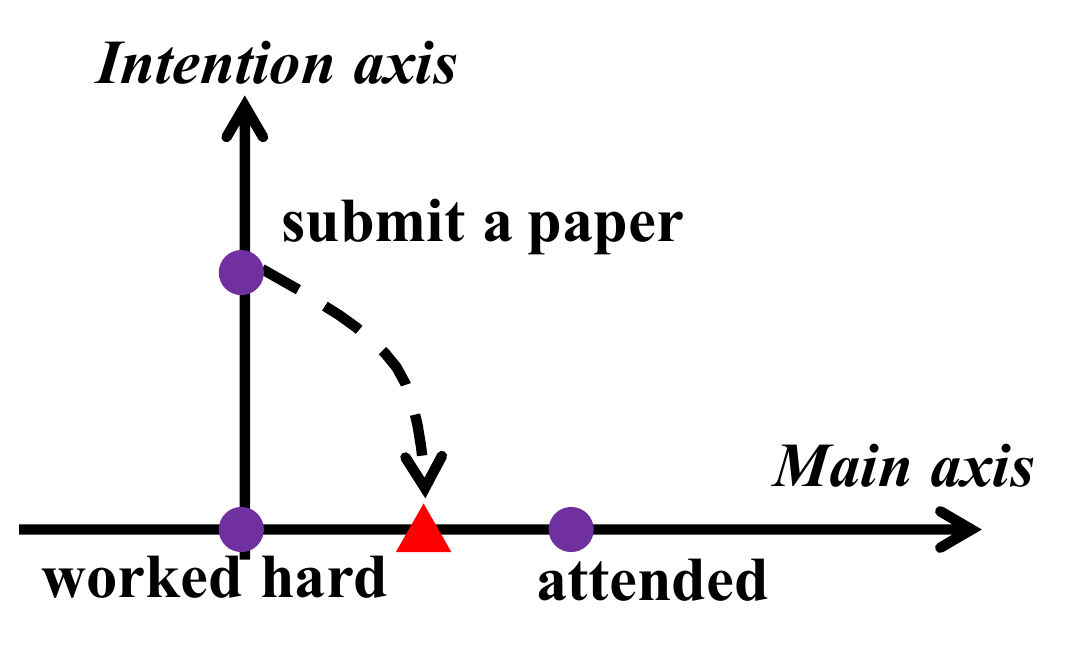}
	\caption{In {\em I worked hard to submit a paper \dots I attended the conference}, the projection of \event{submit a paper} onto the main axis is clearly \rel{before} \event{attended}. However, this projection requires strong external knowledge that a paper should be submitted before attending a conference. Again, this projection is only a guess based on our external knowledge and it is still open whether the paper is submitted or not.}
	\label{fig:projection}
\end{figure}

When strong external knowledge is involved in axis projection, it becomes a reasoning process and the resulting relations are a different type. For example, a reader may reason that in Example~\ref{ex:hard cases 3}, it is well-known that they did ``act again'', implying his \refevent{ev:rebuilding}{rebuilding} had happened and is \rel{before} \refevent{ev:responded}{responded}. Another example is in Fig.~\ref{fig:projection}.
It is obvious that relations based on these projections are not the same with and more challenging than those same-axis relations, so in the current stage, we should focus on same-axis relations only.

\subsubsection{Introduction of the Orthogonal Axes}
\label{subsubsec: orthogonal}
Another prominent difference to earlier work is the introduction of orthogonal axes, which has not been used in any existing work as we know.
A special property is that the intersection event of two axes can be compared to events from both, which can sometimes bridge events, e.g., in Fig.~\ref{fig:multi axes}, \event{Asian crisis} is seemingly \rel{before} \event{hardest hit} due to their connections to \event{expected}.
Since \event{Asian crisis} is on the main axis, it seems that \refevent{ev:hit}{hardest hit} is on the main axis as well.
However, the ``\event{hardest hit}'' in ``\event{Asian crisis} \rel{before} \event{hardest hit}'' is only a projection of the original \refevent{ev:hit}{hardest hit} onto the real axis and is valid only when this \type{Opinion} is true.

Nevertheless, \type{Opinions} are not always true and \type{Intentions} are not always fulfilled.
In Example~\ref{ex:opinion}, \refevent{ev:sponsoring}{sponsoring} and \refevent{ev:resolve}{resolve} are the opinions of the West and the speaker, respectively; whether or not they are true depends on the authors' implications or the readers' understandings, which is often beyond the scope of TempRel annotation.\footnote{For instance, there is undoubtedly a {\em causal} link between \refevent{ev:sponsoring}{sponsoring} and \event{ostracized}.}
Example~\ref{ex:intention} demonstrates a similar situation for \type{Intentions}: when reading the sentence of \refevent{ev:report}{report}, people are inclined to believe that it is fulfilled. But if we read the sentence of \refevent{ev:report2}{report}, we have reason to believe that it is not.
When it comes to \refevent{ev:tell}{tell}, it is unclear if everyone told the truth.
The existence of such examples indicates that orthogonal axes are a better modeling for \type{Intentions} and \type{Opinions}.


\begin{table}[h!]
	\centering\small
	\begin{tabular}{|p{7.5cm}|}
		\hline		
		\textbf{Example~\addexctr\label{ex:opinion}: Opinion events may not always be true.}\\
		\hline
		He is ostracized by the West for (\idxevent{\addeventCtr\label{ev:sponsoring}}{sponsoring}{black}) terrorism.\\
		\hline
		We need to (\idxevent{\addeventCtr\label{ev:resolve}}{resolve}{black}) the deep-seated causes that have resulted in these problems.\\
		\hline
		\textbf{Example~\addexctr\label{ex:intention}: Intentions may not always be fulfilled.}\\
		\hline
		A passerby called the police to (\idxevent{\addeventCtr\label{ev:report}}{report}{black}) the body.\\
		\hline
		A passerby called the police to (\idxevent{\addeventCtr\label{ev:report2}}{report}{black}) the body. Unfortunately, the line was busy.\\
		\hline
		I asked everyone to (\idxevent{\addeventCtr\label{ev:tell}}{tell}{black}) the truth.\\
		\ignore{Police were dispatched to the area to (\idxevent{\addeventCtr\label{ev:restore2}}{restore}{black}) order. However, at least 51 people died in the clashes between police and ethnic Albanians in the region.\\}
		\hline
	\end{tabular}
\end{table}

\subsubsection{Differences from Factuality}
\label{subsubsec:difference to actual}
Event modality have been discussed in many existing event annotation schemes, e.g., Event Nugget \cite{MYHSBKS15}, Rich ERE \cite{SBSRMEWKR15}, and RED.
Generally, an event is classified as \anothertype{Actual} or \anothertype{Non-Actual}, a.k.a. factuality \cite{SauriPu09,LeeArChZe15}.

\QNfinal{The main-axis events defined in this paper seem to be very similar to \anothertype{Actual} events, but with several important differences:}
\textbf{First}, future events are \anothertype{Non-Actual} because they indeed have not happened, but they may be on the main axis.
\textbf{Second}, events that are not on the main axis can also be \anothertype{Actual} events, e.g., intentions that are fulfilled, or opinions that are true.
\textbf{Third}, as demonstrated by Examples~\ref{ex:opinion}-\ref{ex:intention}, identifying anchorability as defined in Table~\ref{tab:nonanchorable} is relatively easy, but judging if an event actually happened is often a high-level understanding task that requires an understanding of the entire document or external knowledge.

Interested readers are referred to Appendix~\ref{appsec:red compare} for a detailed analysis of the difference between \anothertype{Anchorable} (onto the main axis) and \anothertype{Actual} on a subset of RED.

\section{Interval Splitting}
\label{sec:split}

All existing annotation schemes adopt the interval representation of events \cite{Allen84} and there are 13 relations between two intervals (for readers who are not familiar with it, please see Fig.~\ref{fig:13rel} in the appendix).
To reduce the burden of annotators, existing schemes often resort to a reduced set of the 13 relations.
For instance, \citet{VGSHKP07} merged all the overlap relations into a single relation, {\em overlap}.
\citet{BethardMaKl07,DoLuRo12,GormanWrPa16} all adopted this strategy.
In \citet{CassidyMcChBe14}, they further split {\em overlap} into {\em includes}, {\em included} and {\em equal}.

\begin{figure}[htbp!]
	\centering
	\includegraphics[width=0.35\textwidth]{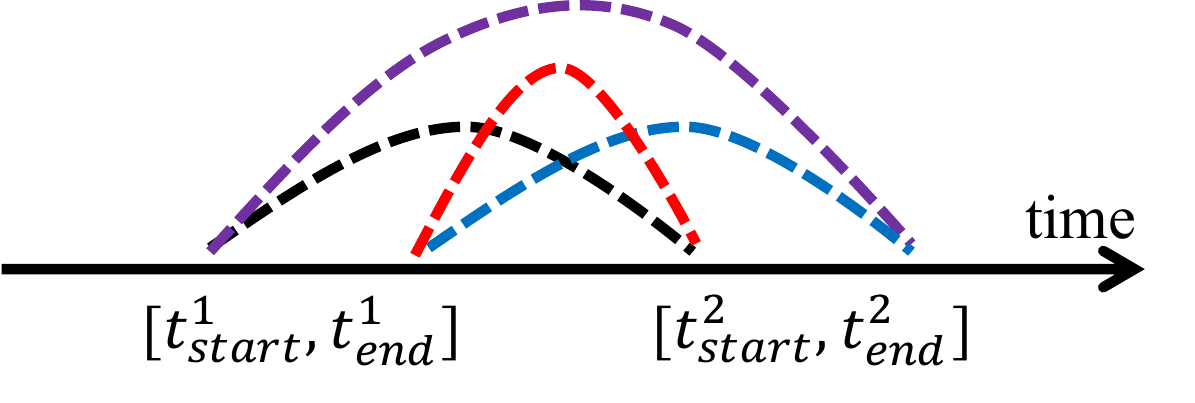}
	\caption{The comparison of two event time intervals, $[t_{start}^1,t_{end}^1]$ and $[t_{start}^2,t_{end}^2]$, can be decomposed into four comparisons， $t_{start}^1$ vs. $t_{start}^2$, $t_{start}^1$ vs. $t_{end}^2$, $t_{end}^1$ vs. $t_{start}^2$, and $t_{end}^1$ vs. $t_{end}^2$, without loss of generality.}
	\label{fig:split}
\end{figure}

Let $[t_{start}^1,t_{end}^1]$ and $[t_{start}^2,t_{end}^2]$ be the time intervals of two events (with the implicit assumption that $t_{start}\le t_{end}$).
Instead of reducing the relations between two intervals, we try to explicitly compare the time points (see Fig.~\ref{fig:split}).
In this way, the label set is simply {\em before, after} and {\em equal},\footnote{We will discuss \rel{vague} in Sec.~\ref{sec:annotation}.} while the expressivity remains the same.
\QNfinal{This interval splitting technique has also been used in \cite{RaFoLa12}.}

In addition to same expressivity, interval splitting can provide even more information when the relation between two events is \rel{vague}. In the conventional setting, imagine that the annotators find that the relation between two events can be either \rel{before} or \rel{before and overlap}. Then the resulting annotation will have to be \rel{vague}, although the annotators actually agree on the relation between $t_{start}^1$ and $t_{start}^2$. Using interval splitting, however, such information can be preserved.

An obvious downside of interval splitting is the increased number of annotations needed (4 point comparisons vs. 1 interval comparison). In practice, however, it is usually much fewer than 4 comparisons. For example, when we see $t_{end}^1<t_{start}^2$ (as in Fig.~\ref{fig:split}), the other three can be skipped because they can all be inferred. Moreover, although the number of annotations is increased, the work load for human annotators may still be the same, because even in the conventional scheme, they still need to think of the relations between start- and end-points before they can make a decision.

\subsection{Ambiguity of End-Points}
\label{subsec:start points}

During our pilot annotation, the annotation quality dropped significantly when the annotators needed to reason about relations involving end-points of events.
Table~\ref{tab:difficulty} shows four metrics of task difficulty when only $t_{start}^1$ vs. $t_{start}^2$ or $t_{end}^1$ vs. $t_{end}^2$ are annotated. Non-anchorable events were removed for both jobs.
The first two metrics, qualifying pass rate and survival rate are related to the two quality control protocols (see Sec.~\ref{subsec:quality} for details). We can see that when annotating the relations between end-points, only one out of ten crowdsourcers (11\%) could successfully pass our qualifying test; and even if they had passed it, half of them (56\%) would have been kicked out in the middle of the task.
The third line is the overall accuracy on gold set from all crowdsourcers (excluding those who did not pass the qualifying test), which drops from 67\% to 37\% when annotating end-end relations.
The last line is the average response time per annotation and we can see that it takes much longer to label an end-end TempRel (52s) than a start-start TempRel (33s).
This important discovery indicates that the TempRels between end-points is probably governed by a different linguistic phenomenon. 

\begin{table}[htbp!]
	\centering\small
	\begin{tabular}{c|c|c}
		\hline
		Metric	&	$t_{start}^1$ vs. $t_{start}^2$	&	$t_{end}^1$ vs. $t_{end}^2$\\
		\hline
		Qualification pass rate	&	50\%	&	11\%\\
		Survival rate			&	74\%	&	56\%\\
		Accuracy on gold		&	67\%	&	37\%\\
		Avg. response time		&	33s		&	52s\\
		\hline
	\end{tabular}
	\caption{Annotations involving the end-points of events are found to be much harder than only comparing the start-points.}
	\label{tab:difficulty}
\end{table}

We hypothesize that the difficulty is a mixture of how durative events are expressed (by authors) and perceived (by readers) in natural language. In cognitive psychology, \citet{CollGe11} discovered that human readers take longer to perceive durative events than punctual events, e.g., {\em owe 50 bucks} vs. {\em lost 50 bucks}. 
From the writer's standpoint, durations are usually fuzzy \cite{SchockaertDe08}, or assumed to be a prior knowledge of readers (e.g., college takes 4 years and watching an NBA game takes a few hours), and thus not always written explicitly.
Given all these reasons, we ignore the comparison of end-points in this work, although event duration is indeed, another important task.




\section{Annotation Scheme Design}
\label{sec:annotation}

To summarize, with the proposed multi-axis modeling (Sec.~\ref{sec:structure}) and interval splitting (Sec.~\ref{sec:split}), our annotation scheme is two-step.
First, we mark every event candidate as being temporally \anothertype{Anchorable} or not (based on the time axis we are working on). Second, we adopt the dense annotation scheme to label TempRels only between \anothertype{Anchorable} events.
Note that we only work on verb events in this paper, so non-verb event candidates are also deleted in a preprocessing step.
\QNfinal{We design crowdsourcing tasks for both steps and as we show later, high crowdsourcing quality was achieved on both tasks.}
In this section, we will discuss some practical issues.

\subsection{Quality Control for Crowdsourcing}
\label{subsec:quality}

We take advantage of the quality control feature in CrowdFlower in our crowdsourcing jobs. For any job, a set of examples are annotated by experts beforehand, which is considered gold and will serve two purposes. (i) Qualifying test: Any crowdsourcer who wants to work on this job has to pass with 70\% accuracy on 10 questions randomly selected from the gold set. (ii) Surviving test: During the annotation process, questions from the gold set will be randomly given to crowdsourcers without notice, and one has to maintain 70\% accuracy on the gold set till the end of the annotation; otherwise, he or she will be forbidden from working on this job anymore and all his/her annotations will be discarded.
At least 5 different annotators are required for every judgement and by default, the majority vote will be the final decision.

\subsection{Vague Relations}
\label{subsec:vague}
How to handle \rel{vague} relations is another issue in temporal annotation. 
In non-dense schemes, annotators usually skip the annotation of a \rel{vague} pair.
In dense schemes, a majority agreement rule is applied as a postprocessing step to back off a decision to \rel{vague} when annotators cannot pass a majority vote \cite{CassidyMcChBe14}, which reminds us that annotators often label a \rel{vague} relation as non-vague due to lack of thinking.

We decide to proactively reduce the possibility of such situations. As mentioned earlier, our label set for $t_{start}^1$ vs. $t_{start}^2$ is \rel{before}, \rel{after}, \rel{equal} and \rel{vague}. We ask two questions: Q1=Is it possible that $t_{start}^1$ is before $t_{start}^2$? Q2=Is it possible that $t_{start}^2$ is before $t_{start}^1$?
Let the answers be A1 and A2. Then we have a one-to-one mapping as follows: A1=A2=yes$\mapsto$\rel{vague}, A1=A2=no$\mapsto$\rel{equal}, A1=yes,~A2=no$\mapsto$\rel{before}, and A1=no,~A2=yes$\mapsto$\rel{after}.
An advantage is that  one will be prompted to think about all possibilities, thus reducing the chance of overlook.

Finally, the annotation interface we used is shown in Appendix~\ref{appsec:interface}.

\section{Corpus Statistics and Quality}
\label{sec:stat}
In this section, we first focus on annotations on the main axis, which is usually the primary storyline and thus has most events. 
Before launching the crowdsourcing tasks, we checked the IAA between two experts on a subset of TB-Dense (about 100 events and 400 relations).
A Cohen's Kappa of .85 was achieved in the first step: anchorability annotation.
Only those events that both experts labeled \anothertype{Anchorable} were kept before they moved onto the second step: relation annotation, for which the Cohen's Kappa was .90 for Q1 and .87 for Q2.
Table~\ref{tab:expert iaa} furthermore shows the distribution, Cohen's Kappa, and F$_1$ of each label. We can see the Kappa and F$_1$ of \rel{vague} ($\kappa$=.75, F$_1$=.81) are generally lower than those of the other labels, confirming that temporal \rel{vagueness} is a more difficult semantic phenomenon.
Nevertheless, the overall IAA shown in Table~\ref{tab:expert iaa} is a significant improvement compared to existing datasets.

\begin{table}[htbp!]
	\centering
	\small
	\begin{tabular}{c|c|c|c|c|c}
		\hline
				&	b	&	a	&	e	&	v	&	Overall\\
		\hline
		Distribution
				&	.49	&	.23	&	.02	&	.26	&	1\\
		IAA: Cohen's $\kappa$
				&	.90	&	.87	&	1	&	.75	&	.84\\
		IAA: F$_1$	&	.92	&	.93	&	1	&	.81	& 	.90\\	
		\hline
	\end{tabular}
	\caption{IAA of two experts' annotations in a pilot study on the main axis. Notations: \textbf{b}efore, \textbf{a}fter, \textbf{e}qual, and \textbf{v}ague.}
	\label{tab:expert iaa}
\end{table}

With the improved IAA confirmed by experts, we sequentially launched the two-step crowdsourcing tasks through CrowdFlower on top of the same 36 documents of TB-Dense.
To evaluate how well the crowdsourcers performed on our task, we calculate two quality metrics: accuracy on the gold set and the Worker Agreement with Aggregate (WAWA).
WAWA indicates the average number of crowdsourcers' responses agreed with the aggregate answer (we used majority aggregation for each question). For example, if $N$ individual responses were obtained in total, and $n$ of them were correct when compared to the aggregate answer, then WAWA is simply $n/N$.
In the first step, crowdsourcers labeled 28\% of the events as \anothertype{Non-Anchorable} to the main axis, with an accuracy on the gold of .86 and a WAWA of .79.

With \anothertype{Non-Anchorable} events filtered, the relation annotation step was launched as another crowdsourcing task.
The label distribution is b=.50, a=.28, e=.03, and v=.19 (consistent with Table~\ref{tab:expert iaa}).
In Table~\ref{tab:stats step2}, we show the annotation quality of this step using accuracy on the gold set and WAWA.
We can see that the crowdsourcers achieved a very good performance on the gold set, indicating that they are consistent with the authors who created the gold set;
these crowdsourcers also achieved a high-level agreement under the WAWA metric, indicating that they are consistent among themselves.
These two metrics indicate that the annotation task is now well-defined and easy to understand even by non-experts.

\begin{table}[htbp!]
	\centering
	\small
	\begin{tabular}{c|c|c|c|c}
		\hline
		No.		&	Metric	&	Q1	&	Q2	&	All\\
		\hline
		1		&	Accuracy on Gold
							&	.89	&	.88	&	.88\\
		2		&	WAWA	&	.82	&	.81	&	.81\\
		\hline
	\end{tabular}
	\caption{Quality analysis of the relation annotation step of \dataset{}. ``Q1'' and ``Q2'' refer to the two questions crowdsourcers were asked (see Sec.~\ref{subsec:vague} for details). Line~1 measures the level of consistency between crowdsourcers and the authors and line~2 measures the level of consistency among the crowdsourcers themselves.}
	\label{tab:stats step2}
\end{table}

\ignore{
Only two gold examples were mis-annotated by crowdsourcers with more than 50\% probability. We show these two sentences in Example~\ref{ex:error analysis 1} and we can see they were indeed difficult ones.

\begin{table}[h!]
	\centering\small
	\begin{tabular}{|p{7.5cm}|}
		\hline
		\textbf{Example~\addexctr\label{ex:error analysis 1}: Errors crowdsourcers made on the gold set in the relation annotation step.}\\
		\hline
		School districts in Omaha and Lincoln, Nebraska, (\idxevent{\addeventCtr\label{ev:called}}{called}{black}) off classes Thursday and city governments (\idxevent{\addeventCtr\label{ev:reported}}{reported}{black}) burning through their entire snow removal budgets with a full two months of winter left.\\
		{\em Gold: \rel{Vague}, i.e., A1=A2=yes}\\
		{\em Crowdsourcers: \rel{Before}, i.e., A1=yes (conf:\missing{86}\%), A2=no (conf:63\%).}\\
		\hline
		They (\idxevent{\addeventCtr\label{ev:arrested}}{arrested}{black}) two men in connection with the bomb factory, which also contained circuitry, detonating cord and the Mitsubishi truck that would have (\idxevent{\addeventCtr\label{ev:carried}}{carried}{black}) the bomb.\\
		\hline
	\end{tabular}
\end{table}
}

We continued to annotate \type{Intention} and \type{Opinion} which create orthogonal branches on the main axis. In the first step, crowdsourcers achieved an accuracy on gold of .82  and a WAWA of .89. 
Since only 16\% of the events are in this category and these axes are usually very short (e.g., {\em \event{allocate} funds to \event{build} a museum}.), the annotation task is relatively small and two experts took the second step and achieved an agreement of .86 (F$_1$).

We name our new dataset {\em \dataset{}} for Multi-Axis Temporal RElations for Start-points.
Each individual judgement cost us \$0.01 and \dataset{} in total cost about \$400 for 36 documents.

\subsection{Comparison to TB-Dense}
To get another checkpoint of the quality of the new dataset, we compare with the annotations of TB-Dense.
TB-Dense has 1.1K verb events, between which 3.4K event-event (EE) relations are annotated.
In the new dataset, 72\% of the events (0.8K) are anchored onto the main axis, resulting in 1.6K EE relations, and 16\% (0.2K) are anchored onto orthogonal axes, resulting in 0.2K EE relations.
The following comparison is based on the 1.8K EE relations in common.
Moreover, since TB-Dense annotations are for intervals instead of start-points only, we converted TB-Dense's interval relations to start-point relations (e.g., if \event{A} includes \event{B}, then $t_{start}^A$ is before $t_{start}^B$).

\begin{table}[htbp!]
	\centering
	\small
	\begin{tabular}{c|c|c|c|c|c}
		\hline
		&	b	&	a	&	e	&	v	&	All\\
		\hline
		b	&	455	&	11	&	5	&	42	&	513\\
		a	&	45	&	309	&	16	&	68	&	438\\
		e	&	13	&	7	&	2	&	10	&	32\\
		v	&	450	&	138	&	20	&	192	&	800\\
		All	&	963	&	465	&	43	&	312	&	1783\\
		\hline
	\end{tabular}
	\caption{An evaluation of \dataset{} against TB-Dense. Horizontal: \dataset{}. Vertical: TB-Dense (with interval relations mapped to start-point relations). Please see explanation of these numbers in text.}
	\label{tab:vs tbdense}
\end{table}

The confusion matrix is shown in Table~\ref{tab:vs tbdense}. A few remarks about how to understand it:
\textbf{First}, when TB-Dense labels \rel{before} or \rel{after}, \dataset{} also has a high-probability of having the same label (b=455/513=.89, a=309/438=.71); when \dataset{} labels \rel{vague}, TB-Dense is also very likely to label \rel{vague} (v=192/312=.62). This indicates the {\em high agreement level} between the two datasets if the interval- or point-based annotation difference is ruled out.
\textbf{Second}, many \rel{vague} relations in TB-Dense are labeled as \rel{before}, \rel{after} or \rel{equal} in \dataset{}. This is expected because TB-Dense annotates relations between {\em intervals}, while \dataset{} annotates {\em start-points}.
When durative events are involved, the problem usually becomes more difficult and interval-based annotation is more likely to label \rel{vague} (see earlier discussions in Sec.~\ref{sec:split}). 
Example~\ref{ex:error analysis tbdense} shows three typical cases, where \refevent{ev:became}{became}, \refevent{ev:backed}{backed}, \refevent{ev:rose}{rose} and \refevent{ev:extending}{extending} can be considered durative. 
If only their start-points are considered, the crowdsourcers were correct in labeling \event{e\ref{ev:became}} before \event{e\ref{ev:said2}}, \event{e\ref{ev:discussed}} after \event{e\ref{ev:backed}}, and \event{e\ref{ev:rose}} equal to \event{e\ref{ev:extending}}, although TB-Dense says \rel{vague} for all of them.
\textbf{Third}, \rel{equal} seems to be the relation that the two dataset mostly disagree on, which is probably due to crowdsourcers' lack of understanding in time granularity and event coreference. Although \rel{equal} relations only constitutes a small portion in all relations, it needs further investigation.

\begin{table}[h!]
	\centering\small
	\begin{tabular}{|p{7.5cm}|}
		\hline
		\textbf{Example~\addexctr\label{ex:error analysis tbdense}: Typical cases that TB-Dense annotated \rel{vague} but \dataset{} annotated \rel{before}, \rel{after}, and \rel{equal}, respectively.}\\
		\hline
		At one point , when it (\idxevent{\addeventCtr\label{ev:became}}{became}{black}) clear controllers could not contact the plane, someone (\idxevent{\addeventCtr\label{ev:said2}}{said}{black}) a prayer.\\
		{\em TB-Dense: \rel{vague}; \dataset{}: \rel{before}}\\
		\hline
		The US is bolstering its military presence in the gulf, as President Clinton (\idxevent{\addeventCtr\label{ev:discussed}}{discussed}{black}) the Iraq crisis with the one ally who has (\idxevent{\addeventCtr\label{ev:backed}}{backed}{black}) his threat of force, British prime minister Tony Blair.\\
		{\em TB-Dense: \rel{vague}; \dataset{}: \rel{after}}\\
		\hline
		Average hourly earnings of nonsupervisory employees (\idxevent{\addeventCtr\label{ev:rose}}{rose}{black}) to \$12.51.
		The gain left wages 3.8 percent higher than a year earlier, (\idxevent{\addeventCtr\label{ev:extending}}{extending}{black}) a trend that has given back to workers some of the earning power they lost to inflation in the last decade.\\
		{\em TB-Dense: \rel{vague}; \dataset{}: \rel{equal}}\\
		\hline
		\ignore{
		A new task force began delving Thursday into the slayings of 14 black women in the Newark area, as law-enforcement officials (\idxevent{\addeventCtr\label{ev:acknowledged}}{acknowledged}{black}) that they needed to work harder to solve the cases of murdered women. The police and prosecutors (\idxevent{\addeventCtr\label{ev:said3}}{said}{black}) they had ... yet to find any pattern linking the killings or the victims...\\
		{\em TB-Dense: \rel{vague}; \dataset{}: \rel{equal}}\\
		\hline}
	\end{tabular}
\end{table}

\section{Baseline System}
\label{sec:result}
We develop a baseline system for TempRel extraction on \dataset{}, assuming that all the events and axes are given.
The following commonly-used features for each event pair are used: (i) The part-of-speech (POS) tags of each individual event and of its neighboring three words. (ii) The sentence and token distance between the two events. (iii) The appearance of any modal verb between the two event mentions in text (i.e., \textsf{will, would, can, could, may} and \textsf{might}). (iv) The appearance of any temporal connectives between the two event mentions (e.g., \textsf{before, after} and \textsf{since}). (v) Whether the two verbs have a common synonym from their synsets in WordNet \cite{Fellbaum98}. (vi) Whether the input event mentions have a common derivational form derived from WordNet. (vii) The head words of the preposition phrases that cover each event, respectively. And (viii) event properties such as Aspect, Modality, and Polarity \QNfinal{that come with the TimeBank dataset and are commonly used as features}.

The proposed baseline system uses the averaged perceptron algorithm to classify the relation between each event pair into one of the four relation types. We adopted the same train/dev/test split of TB-Dense, where there are 22 documents in train, 5 in dev, and 9 in test. 
Parameters were tuned on the train-set to maximize its F$_1$ on the dev-set, after which the classifier was retrained on the union of train and dev.
A detailed analysis of the baseline system is provided in Table~\ref{tab:baseline}. 
The performance on \rel{equal} and \rel{vague} is lower than on \rel{before} and \rel{after}, probably due to shortage in these labels in the training data and the inherent difficulty in event coreference and temporal vagueness.
We can see, though, that the overall performance on \dataset{} is much better than those in the literature for TempRel extraction, which used to be in the low 50's \cite{ChambersCaMcBe14,NingFeRo17}.
The same system was also retrained and tested on the original annotations of TB-Dense (Line ``Original''), which confirms the significant improvement if the proposed annotation scheme is used.
Note that we {\em do not} mean to say that the proposed baseline system itself is better than other existing algorithms, but rather that the proposed annotation scheme and the resulting dataset lead to better defined machine learning tasks.
In the future, more data can be collected and used with advanced techniques such as ILP \cite{DoLuRo12}, structured learning \cite{NingFeRo17} or multi-sieve \cite{ChambersCaMcBe14}.

\begin{table}[htbp!]
\centering
\small
\begin{tabular}{c|c|c|c|c|c|c}
	\hline
		&	\multicolumn{3}{c|}{Training}	&	\multicolumn{3}{c}{Testing}\\
	\cline{2-7}
		&	P	&	R	&	F$_1$	&	P	&	R	&	F$_1$\\
	\hline
	Before	&	.74	&	.91	&	.82		&	.71	&	.80	&	.75	\\
	After	&	.73	&	.77	&	.75		&	.55	&	.64	&	.59\\
	Equal	&	1	&	.05	&	.09		&	-	&	-	&	-\\
	Vague	&	.75	&	.28	&	.41		&	.29	&	.13	&	.18\\
	\hline
	Overall	&	\best{.73}	&	\best{.81}	&	\best{.77}		&	\best{.66}	&	\best{.72}	&	\best{.69}\\
	\hline\hline
	Original&	.44	&	.67	&	.53		&	.40	&	.60	&	.48\\
	\hline
\end{tabular}
\caption{Performance of the proposed baseline system on \dataset{}. Line ``Original'' is the same system retrained on the original TB-Dense and tested on the same subset of event pairs. Due to the limited number of \rel{equal} examples, the system did not make any \rel{equal} predictions on the testset.}
\label{tab:baseline}
\end{table}

\section{Conclusion}
\label{sec:conclusion}

This paper proposes a new scheme for TempRel annotation between events, simplifying the task by focusing on a single time axis at a time.
We have also identified that end-points of events is a major source of confusion during annotation due to reasons beyond the scope of TempRel annotation, and proposed to focus on start-points only and handle the end-points issue in further investigation (e.g., in event duration annotation tasks).
Pilot study by expert annotators shows significant IAA improvements compared to literature values, indicating a better task definition under the proposed scheme. This further enables the usage of crowdsourcing to collect a new dataset, \dataset{}, at a lower time cost. 
Analysis shows that \dataset{}, albeit crowdsourced, has achieved a reasonably good agreement level, as confirmed by its performance on the gold set (agreement with the authors), the WAWA metric (agreement with the crowdsourcers themselves), and consistency with TB-Dense (agreement with an existing dataset).
Given the fact that existing schemes suffer from low IAAs and lack of data, we hope that the findings in this work would provide a good start towards understanding more sophisticated semantic phenomena in this area.

\ignore{
\missing{Clinical temporal relation datasets}

\missing{Quadratic issue still unsolved.}
Still, one important issue remain unsolved is the quadratic nature of the number of event pairs in terms of the number of events. Even crowdsourcing cannot afford to label all pairs of events. We may need learning approaches that can get supervision from partially annotated graphs. The scheme proposed by EventTimeCorpus \cite{ReimersDeGu16} may also be a good complementary.

\missing{solving event filtering and tlinks jointly}
}

\section*{Acknowledgements}
\QNfinal{We thank Martha Palmer, Tim O'Gorman, Mark Sammons and all the anonymous reviewers for providing insightful comments and critique in earlier stages of this work. This research is supported in part by a grant from the Allen Institute for Artificial Intelligence (allenai.org); the IBM-ILLINOIS Center for Cognitive Computing Systems Research (C3SR) - a research collaboration as part of the IBM AI Horizons Network; by DARPA under agreement number FA8750-13-2-0008; and by the Army Research Laboratory (ARL) under agreement W911NF-09-2-0053 (the ARL Network Science CTA).}

\QNfinal{The U.S. Government is authorized to reproduce and distribute reprints for Governmental purposes notwithstanding any copyright notation thereon. 
The views and conclusions contained herein are those of the authors and should not be interpreted as necessarily representing the official policies or endorsements, either expressed or implied, of DARPA, of the Army Research Laboratory or the U.S. Government.
Any opinions, findings, conclusions or recommendations are those of the authors and do not necessarily reflect the view of the ARL.}

\bibliography{acl2018,kbcom18,cited-long,ccg-long}
\bibliographystyle{acl_natbib}

\newpage\onecolumn
\appendix
\section{Examples of Table~\ref{tab:nonanchorable}}
\label{appsec:definition}
The names of those categories in Table~\ref{tab:nonanchorable} are straightforward. Here we further provide examples for each of them in Example~\ref{ex:definition}. Note that most of them are consistent with the definitions in the literature, with one exception for \type{Intention}. In TimeML \cite{Pustejovsky03}, there are two types of intentions, I-Action (e.g., {\em attempt, try} and {\em promise}) and I-State (e.g., {\em believe, intend} and {\em want}). But our definition of intention is the actual intent of these verbs. For example, in Example~\ref{ex:definition}, \event{e\ref{ev:leave}} and \event{e\ref{ev:build2}} are \type{Intention}. This definition is more general so that verbs that are not I-Action or I-State can still create orthogonal axis of intention, e.g., the verb ``allocated'' in the sentence of \event{e\ref{ev:build2}}.
\begin{table}[h]
	\centering\small
	\begin{tabular}{|p{7.5cm}|}
		\hline
		\textbf{Example~\addexctr\label{ex:definition}}\\
		\hline
		\textbf{[Orthogonal axis] \type{Intention}/\type{Opinion}}\\
		I plan/want to (\idxevent{\addeventCtr\label{ev:leave}}{leave}{black}) tomorrow. \\
		The mayor has allocated funds to (\idxevent{\addeventCtr\label{ev:build2}}{build}{black}) a museum.\\
		I think he will (\idxevent{\addeventCtr\label{ev:win}}{win}{black}) the race.\\
		\hline
		\textbf{[Parallel axis] \type{Hypothesis}/\type{Generic}}\\
		If I'm (\idxevent{\addeventCtr\label{ev:elected}}{elected}{black}), I will cut income tax. \\
		If I'm elected, I will (\idxevent{\addeventCtr\label{ev:cut}}{cut}{black}) income tax.\\
		Fruit (\idxevent{\addeventCtr\label{ev:contains}}{contains}{black}) water.\\
		Lions (\idxevent{\addeventCtr\label{ev:hunt2}}{hunt}{black}) zebras.\\
		\hline
		\textbf{[Not on any axis] \type{Negation}}\\
		The financial assistance from the Wolrd Bank is not (\idxevent{\addeventCtr\label{ev:helping}}{helping}{black}). \\
		They don't (\idxevent{\addeventCtr\label{ev:want}}{want}{black}) to play with us.\\
		He failed to (\idxevent{\addeventCtr\label{ev:find}}{find}{black}) buyers.\\
		\hline
		\textbf{[Other] \type{Static}/\type{Recurrent}}\\
		He (\idxevent{\addeventCtr\label{ev:is}}{is}{black}) brave. \\
		New York (\idxevent{\addeventCtr\label{ev:is2}}{is}{black}) on the east coast.\\
		The shuttle will be (\idxevent{\addeventCtr\label{ev:departing}}{departing}{black}) at 6:30am every day.\\
		\hline
	\end{tabular}
\end{table}

\begin{figure}[htbp!]
	\centering
	\includegraphics[width=0.45\textwidth]{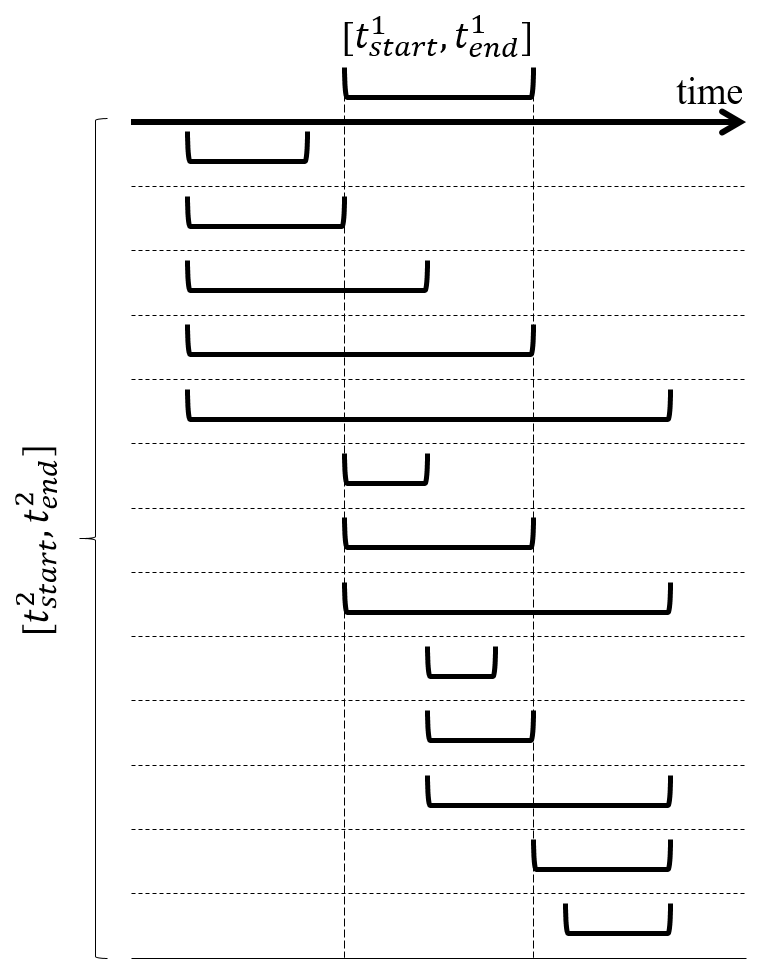}
	\caption{Thirteen possible relations between two events whose timescopes are $[t_{start}^1,t_{end}^1]$ and $[t_{start}^2,t_{end}^2]$ (from top to bottom): {\em after, immediately after, after and overlap, ends, included, started by, equal, starts, includes, ended by, before and overlap, immediately before} and {\em before}.}
	\label{fig:13rel}
\end{figure}

\section{\anothertype{Anchorable} vs. \anothertype{Actual}}
\label{appsec:red compare}
As discussed in the paper, when we check if an event is \anothertype{Anchorable} onto the main axis, it seems very similar to annotating whether an event is \anothertype{Actual} in REALIS labeling.
We have discussed the differences in Sec.~\ref{subsubsec:difference to actual}.
To better understand them, we randomly selected 5 documents from RED \cite{GormanWrPa16}, where there are 314 events, 166 of which are verbs (we only handle verb events).
Two experts annotated the anchorability of these 166 verb events independently without looking at the original REALIS annotation from RED, and they achieved a Cohen's Kappa of .88 in anchorability annotation, consistent with their Cohen's Kappa achieved on \dataset{}.
To aggregate the result from two experts, we mark an event as \anothertype{Anchorable} only when both experts labeled \anothertype{Anchorable}.
As for REALIS labeling in RED, we group \type{Generic}, \type{Hypothetical}, and \type{Hedged} into a single label of \anothertype{Non-Actual}.

\begin{table}[htbp!]
	\centering
	\begin{tabular}{c|c|c|c}
		\hline
		\multicolumn{2}{c|}{}&\multicolumn{2}{c}{\anothertype{Anchorable}}\\
		\cline{3-4}
		\multicolumn{2}{c|}{}&Yes&No\\
		\hline
		\multirow{2}{*}{\anothertype{Actual}}&Yes&108&25\\\cline{2-4}&No&0&33\\
		\hline
	\end{tabular}
	\caption{Comparison between anchrability and factuality on a subset of verb events randomly selected from RED.}
	\label{tab:versus red}
\end{table}

The comparison between \anothertype{Anchorable} and \anothertype{Actual} is shown in Table~\ref{tab:versus red}. On this subset of 166 events, we did not see \anothertype{Anchorable} events that are \anothertype{Non-Actual} because such cases are indeed less frequent in practice; the only difference is that we annotated 25 events as \anothertype{Non-Anchorable}, while RED annotated them as \anothertype{Actual}.
Among the 25 different cases, 11 are \type{Intention}, 4 are \type{Opinion}, 6 are \type{Static}, and 4 are \type{Negation}.
Typical examples from each category are shown in Example~\ref{ex:versus red}.
Note that if we calculate the McNemar's statistics based on Table~\ref{tab:versus red}, \anothertype{Anchorable} and \anothertype{Actual} are statistically different with $p\ll0.001$.

\begin{table}[htbp!]
	\centering\small
	\begin{tabular}{|p{7.5cm}|}
		\hline
		\textbf{Example~\addexctr\label{ex:versus red}: Typical cases that RED annotated \anothertype{Actual} and we annotated \anothertype{Non-Anchorable}.}\\
		\hline
		Libya has since agreed to (\idxevent{\addeventCtr\label{ev:pay}}{pay}{black}) compensation to the families of the Berlin disco victims as well as the families of the victims of the 1988 Pan Am 103 bombing over Lockerbie, Scotland, which killed 270 people, including 189 Americans. [\textbf{We think it is \type{Intention}}]\\
		\hline
		Gadhafi had long been ostracized by the West for (\idxevent{\addeventCtr\label{ev:sponsoring2}}{sponsoring}{black}) terrorism, but in recent years sought to emerge from his pariah status by abandoning weapons of mass destruction and renouncing terrorism in 2003. [\textbf{We think it is \type{Opinion}}]\\
		\hline
		We need to resolve the deep-seated causes that have resulted in these problems,  Premier Wen said in an interview with Hong Kong-(\idxevent{\addeventCtr\label{ev:based}}{based}{black}) Phoenix Television. [\textbf{We think it is \type{Static}}]\\
		\hline
		Fuel prices had been frozen for six years, but the government said it could no longer afford to (\idxevent{\addeventCtr\label{ev:subsidize}}{subsidize}{black}) them. [\textbf{We think it is \type{Negation}}]\\
		\hline
	\end{tabular}
\end{table}

\section{Annotation Interface}
\label{appsec:interface}

\begin{figure*}[htbp!]
	\centering
	\includegraphics[width=0.7\textwidth]{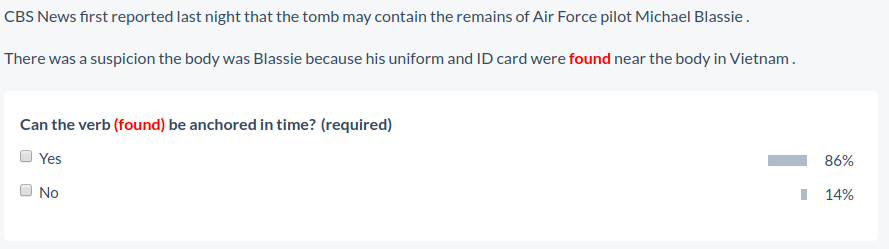}
	\caption{Annotation interface for the first step: temporal anchorability. The owner of the task can see the crowdsourcers' distribution of each answer (e.g., 86\% and 14\%), which is of course not available to crowdsourcers.}
	\label{fig:gui-anchor}
\end{figure*}

The annotation interface was designed based on the web interface of CrowdFlower.
In the anchorability annotation step (i.e., the first step), we show each crowdsourcer one event at a time, along with the full context of this event. Crowdsourcers only need to make a binary decision of Yes/No, as shown in Fig.~\ref{fig:gui-anchor}.

\begin{figure*}[htbp!]
	\centering
	\includegraphics[width=0.65\textwidth]{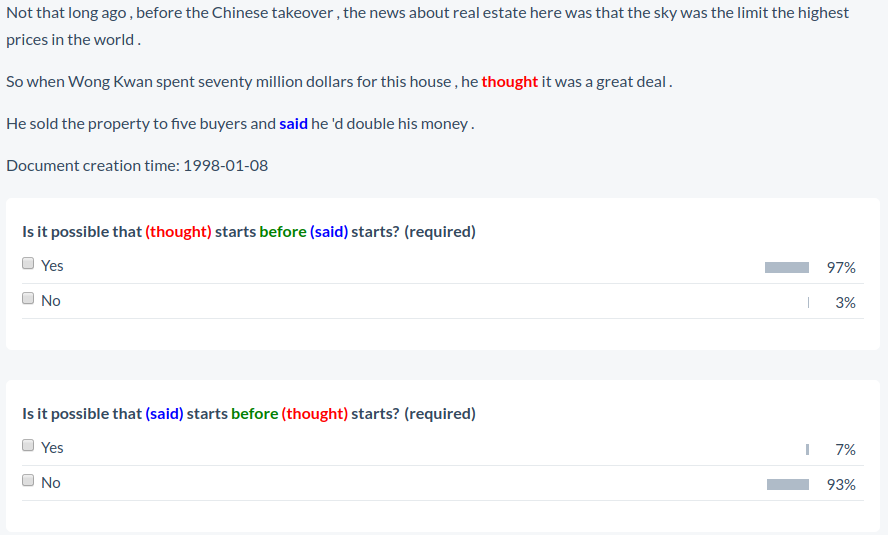}
	\caption{Tentative annotation interface for the second step: relation annotation. This design gives crowdsourcers the wrong impression to select one ``yes'' and one ``no'' for Q1 and Q2, leading to strong correlation between answers of Q1 and answers of Q2.}
	\label{fig:gui-relation old}
\end{figure*}

\begin{figure*}[htbp!]
	\centering
	\begin{subfigure}[b]{\textwidth}
		\centering
		\includegraphics[width=0.65\textwidth]{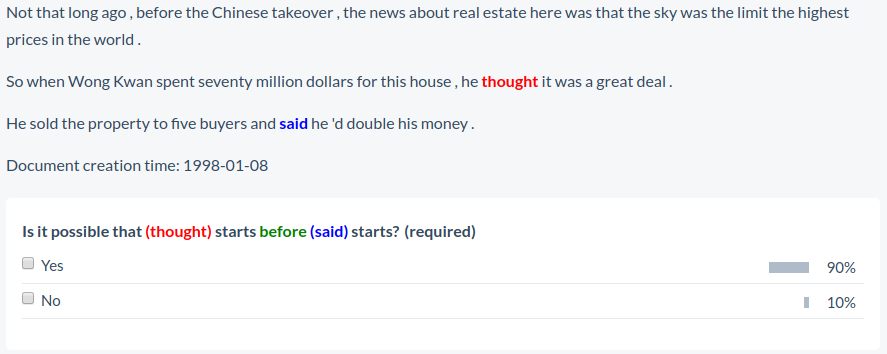}
		\caption{Task A: Only ask Q1}
	\end{subfigure}%
	\\
	\begin{subfigure}[b]{\textwidth}
		\centering
		\includegraphics[width=0.65\textwidth]{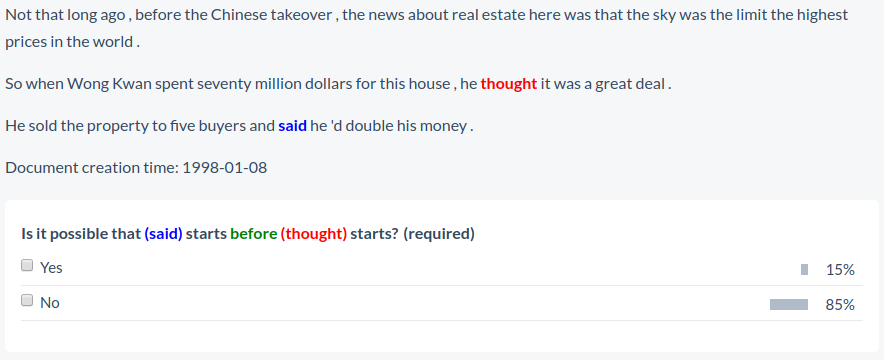}
		\caption{Task B: Only ask Q2}
	\end{subfigure}
	\caption{The final annotation interface, where Q1 and Q2 are posed in separate tasks so that a single annotator will not see both two questions simultaneously, forcing them to think the temporal relation carefully instead of simply putting the opposite answer to the other question.}
	\label{fig:gui-relation}
\end{figure*}

The interface design for the relation annotation step (i.e., the second step) is tricky.
As explained in Sec.~\ref{subsec:vague}, we need to ask two questions for each pair of events to figure out the actual TempRel:
Q1=Is it possible that $t_{start}^1$ is before $t_{start}^2$? Q2=Is it possible that $t_{start}^2$ is before $t_{start}^1$?
We notice in practice that asking Q1 and Q2 simultaneously (as shown in Fig.~\ref{fig:gui-relation old}) gives annotators the wrong impression that there has to be one ``yes'' and one ``no''.
Therefore, we decide to ask Q1 and Q2 separately. Specifically, we launch two separate tasks. One task only has Q1 (Task A), and the other only has Q2 (Task B), so that a same annotator is guaranteed not to see Q1 and Q2 simultaneously (as shown in Fig.~\ref{fig:gui-relation}).

\end{document}